\documentclass[reqno ,11pt]{article}
\usepackage[margin = 1in]{geometry}
 \usepackage{bibentry}
\usepackage{placeins}
\usepackage[space]{grffile}
\usepackage{amssymb}
\usepackage{epstopdf}
\usepackage{fullpage}
\usepackage{pslatex}
\usepackage{amsfonts}
\usepackage{setspace}
\usepackage{float}
\usepackage{fancyhdr}
\usepackage{amsmath,amssymb,amsthm,amsfonts}
\usepackage{soul} 
\usepackage{array}
\usepackage{empheq}
\usepackage{enumerate}
\usepackage{graphics,graphicx}
\usepackage{longtable}
\usepackage{xfrac}
\usepackage{minibox}
\usepackage{caption}
\usepackage{subcaption}
\usepackage{verbatim}
\usepackage{listings}
\usepackage{lstautogobble}
\usepackage{hyperref}
\usepackage[utf8]{inputenc}
\usepackage[T1]{fontenc}
\usepackage{tikz}
\usetikzlibrary{positioning,shapes,shadows,arrows,calc}
\tikzstyle{line}=[draw, very thick, black]
\tikzstyle{auxline}=[draw, very thin, black]
\tikzstyle{arrow}=[draw, -latex, very thick, black]
\usetikzlibrary{fit}
\usepackage{chngcntr}
\usepackage{multicol}
\usepackage{makecell}
\usepackage{authblk}
\usepackage{booktabs}
\usepackage[ruled,vlined]{algorithm2e}

\DeclareGraphicsExtensions{.pdf}
\begin{document}

\title{Classification Models for Partially Ordered Sequences} 
\author{Stephanie Ger$^1$, Diego Klabjan $^2$, Jean Utke $^3$}
\date
{$^1$ Department of Engineering Sciences and Applied Mathematics, Northwestern University, Evanston, Illinois, USA \\
$^2$ Department of Industrial Engineering and Management Sciences, Northwestern University, Evanston, Illinois, USA\\
$^3$ Allstate Insurance Company\\
\today}
\maketitle

\begin{abstract}
\noindent Many models such as Long Short Term Memory (LSTMs), Gated Recurrent Units (GRUs) and transformers have been developed to classify time series data with the assumption that events in a sequence are ordered. On the other hand, fewer models have been developed for set based inputs, where order does not matter. There are several use cases where data is given as partially-ordered sequences because of the granularity or uncertainty of time stamps. We introduce a novel transformer based model for such prediction tasks, and benchmark against extensions of existing order invariant models. We also discuss how transition probabilities between events in a sequence can be used to improve model performance. We show that the transformer-based equal-time model outperforms extensions of existing set models on three data sets. 
\end{abstract}

\section{Introduction}
With the development of Recurrent Neural Networks (RNNs), many model architectures such as LSTMs and GRUs have been used to classify time series data \cite{LSTM, GRU}. Extensions such as the attention mechanism have improved classification accuracy dramatically \cite{attention2015}. Attention seeks to improve model performance by learning a trainable weighting for the relative importance of model inputs. Other improvements include architectures such as sequence-to-sequence or sequence-to-one which can be used for video captioning or sentiment analysis, respectively. In addition, attention based models such as transformers have been developed for sequence classification \cite{transformer}. 

However, fewer models have been built for set, or order-invariant, inputs. Examples of order-invariant data sets include estimating the red shift of a cluster of galaxies \cite{deep-sets}. Existing set-based classification models seek to classify sets by building an order-invariant layer. These models do so by either summing a representation of the inputs or by using attention to determine an order-invariant representation \cite{order-matters, deep-sets}. 

With the advent of Internet of Things (IoT) and sensor data, it is possible to have inputs from multiple sensors where the order of the inputs is unknown. That is, the time granularity may not be fine enough, leading to multiple inputs, or events, at the same time step. The na\"{i}ve approach is to average all equal-time events, or time steps with multiple events, and then apply standard sequence models to the output. This is problematic as information is lost through averaging. We propose models that first generate a single representation for equal-time events and then use that representation along with the remaining events in the sequence as the input to a classifier such as a transformer or LSTM. We also propose a model that uses the transition matrix, computed using the subset of ordered events in the training set, as an input to the equal-time layer to determine a single representation that attempts to better capture the true ordering of the data. We propose novel transformer-based methods with and without the transition matrix input and benchmark against extensions of existing set-based models. The set-based transformer models significantly improve on the existing models by 7.4\% to 10.1\% depending on the data set. Furthermore, we observe an 8.71\% improvement over the set-based transformer model when the transition matrix input is used. The main contributions are as follows:

\begin{enumerate}
\item a novel transformer-based model for representing equal-time events for partially ordered time series data so that standard sequence classification models can be applied; 
\item a novel model that uses a transition matrix to order equal-time events that improves model performance;
\item a computational benchmarking study of existing set models on the partially-ordered data sets. 
\end{enumerate}
In the next section, we discuss relevant literature. Section 3 discusses all of the models, while the computational results are presented in Section 4.

\section{Literature Review}

Recently, some models have been developed to deal with unordered set data. These include models that build order-invariant neural networks as well as models that aim to order a set of data. The order-invariant models can be used for tasks such as adding a set of numbers together or sorting a set. Models for set ordering have been developed primarily for ordering a bag of words into a semantically correct sentence. 

In \cite{deep-sets}, a definition of order invariant functions on a set is provided, and it is shown that it is possible to build a neural network architecture that is order-invariant. The Deep Sets model achieves order invariance by applying the same dense layer to each input and summing the outputs.  It is suited for tasks such as summing a set of numbers together. Another method defines a way to use an attention mechanism to build a recurrent model that is order-invariant \cite{order-matters}. With recurrent models that are invariant to an input set order, one can build set-to-set, set-to-sequence and sequence-to-set models. In particular, set-to-sequence models can be used to impose an order on an input set. 

Many models have been developed that take a set of words as input and produce a semantically correct sentence as output, for example, $n$-gram language models and Statistical Machine Translation \cite{n-gram, SMT}. In \cite{word-ordering}, the model orders a bag of words by using an LSTM hidden state to estimate the probability that a word occurs next in the sentence. These probabilities are then used in adaptive beam search to determine the highest probability ordering of words. Thus, the loss function takes into account both the conditional probability the current word is next given the words that have already been ordered as well as an estimate of the yet-to-be-ordered words. 

However, these models are designed for data that is entirely unordered, not partially ordered sequence data. With partially ordered sequence data, it is possible to determine transition probabilities between events from the ordered events, and to use that to order the unordered events in the data assuming said data has an unknown inherent ordering. The idea of using a transition matrix for unordered events borrows from Hidden Markov Models which can learn the most likely sequence of events by considering transition probabilities \cite{HMM}.

Transformers are an attention based model architecture that have been used for encoder-decoder sequence models. Instead of an LSTM cell, the attention mechanism is used to determine the output of the encoder and decoder, by considering the interaction between an event in a sequence and all other events in the sequence. As the attention is computed between each event and the remaining events in the sequence, it seems reasonable that attention used in transformer-based models can also be applied to ordering sequential data where the order is unknown \cite{transformer}. The difference between the problem that we consider and the set-based models is that we want to determine a representation for the unordered data for use within the context of a sequence model, while set-based models aim at order invariance. Therefore, as transformer models are built on the basis of examining the relative importance of each event in the sequence, it makes sense to base partially-ordered sequence models on transformers.

Transformers have been used in language models such as BERT and XLNet \cite{bert, xlnet}. In particular, the XLNet model has an argument in the transformer self-attention, where a relative segment encoding for elements in the sequence is used to compute the attention. The segment encoding value varies based on elements being from the same or a different segment. We borrow this concept to capture a transition matrix input in an equal-time model. 

\section{Approaches}
We consider a partially-ordered sequence $x = (x_1, x_2, \dots, x_T) \in \mathcal{X}$ where each element $x_i$ of the sequence consists of up to $N$ unordered events, each represented by a vector of dimension $M$; we have $x_i = \{e_1, e_2, \dots, e_n\}$ where $e_j \in R^M$ and $n \le N$. Sequence length $T$ can vary by sequence. In this section, we discuss equal-time models for determining a representation $\tilde{x_i} \in R^M$ for all equal-time events in the sequence. Note that whenever $n = 1$, we set $\tilde{x_i} = x_i$. Each of these set representation models is then incorporated in either an LSTM or a transformer model on the sequence $\tilde{x} = (\tilde{x}_1, \dots, \tilde{x_i}, \dots, \tilde{x}_T)$.

\subsection{Deep Sets}
For the deep set model, the same dense layer is applied to the $n$ unordered events and then the outputs are summed to get \[\tilde{x_i} = \frac1n\sum_{i\in n} \mathbf{W}e_i\] where $W$ are trainable weights. As discussed in \cite{deep-sets}, this model is invariant to the order of inputs as the parameters are shared and by applying a dense layer we expect the model to learn relevant features for each of the unordered events.

\subsection{LSTM Set} 
We consider the LSTM and attention mechanism technique for order invariant models to compute a single representation for unordered equal-time events \cite{order-matters}. To generate a representation for the equal-time events $x_i = \{e_1, e_2, \dots, e_n\}$, we apply attention and the LSTM equations $N$ times and use the final output of the LSTM equation as the representation $\tilde{x}_i$ for the unordered events. By applying the attention repeatedly, we compute a representation that encodes the relative importance of each equal-time event. The cell and hidden state ($q_0$ and $r_0$) are both initialized as a constant vector and the same attention and LSTM weights are applied at each of the $N$ iterations. We apply the following equations for each of the $N$ iterations and use the final $q_t^*$ as the representation for $x_i$ where $q_t^*$ is determined by concatenating $q_t$ and $r_t$. 

\begin{equation}
\begin{aligned}\label{eq:1}
q_t &= LSTM(q_{t-1}^*)\\ 
d_{i,t} &= V\cdot \text{tanh}(W\cdot [e_i, q_t])\\ 
a_{i,t} &= \frac{\exp(d_{i,t})}{\sum_j \exp(d_{j,t})} \\ 
q_t^* &= [q_t, r_t]
\end{aligned}
\end{equation}

\subsection{Transformers for Equal-Time Events}
As in the standard transformer model, we use the set of equal-time events $ \{e_1, e_2, \dots, e_n\}$ and consider weights $W_K, W_V$ and $K_Q$ to compute the value, key and query vectors to compute self-attention. For each attention head, we compute self-attention to get the output of the attention vectors. As the transformer output is the same length as the transformer input, we use attention to output a single representation for each equal-time event. The number of attention heads is a tunable hyperparameter. The difference with standard transformers is that we do not employ positional encoding. 

\subsubsection{Transition Matrix Input for Transformers}
For a dataset where we have inherently ordered data, we can use the ordered events in the training data to compute a transition matrix $T$. Instead of using the transition matrix, which represents only probabilities, to order the unordered equal-time events we alter the model such that the transition matrix can be used in the attention computation of the transformer. Furthermore, the transition matrix might not imply a total order. For example, with a text based data set, one could compute transition probabilities between different parts-of-speech or between different words, though the latter transition matrix is not likely to be useful for computing a total order. Given $T$ and the transformer model discussed earlier, we alter the computation for each attention head so that given the query, value and key vector denoted by $q_i$, $k_i$ and $v_i$ respectively, instead of the standard transformer attention equation \[x_j = \sum_i \text{softmax}(q_jk_i)v_i,\] we consider the transformer attention equation \[x_j = \sum_i\left[\text{softmax}(q_jk_i) + (q_j + b)^TT_{ij}\right]v_i.\] Here, $T_{ij}$ is the transition probability from event $e_i$ to event $e_j$ for $e_i, e_j \in x_i$ where $x_i$ is an equal-time event and $b$ is a trainable bias.

\section{Computational Study}
We consider four sequential datasets\footnote{Code and data are available at \href{https://github.com/stephanieger/equal-time/}{\url{https://github.com/stephanieger/equal-time/}}. Three datasets are provided.} . Each of these datasets consists of multi-feature sequence data where a subset of time steps in the sequence contains a set of unordered events. The first dataset is sensor data from environmental sensors placed around the city of Chicago where we predict if the sensor reading values exceed a threshold in the next 12 hours \footnote{\href{https://aot-file-browser.plenar.io/}{\url{https://aot-file-browser.plenar.io/}}}. The second dataset consists of power readings from household sensors where we predict if five appliances are turned on or off based on power usage \footnote{\href{https://dataverse.harvard.edu/dataset.xhtml?persistentId=doi:10.7910/DVN/FIE0S4}{\url{https://dataverse.harvard.edu/dataset.xhtml?persistentId=doi:10.7910/DVN/FIE0S4}}} \cite{AMP}. The third dataset considers a meeting transcript dataset where we have multiple speakers who may be speaking at the same time \footnote{\href{http://groups.inf.ed.ac.uk/ami/download/}{\url{http://groups.inf.ed.ac.uk/ami/download/}}}\cite{AMI}. The goal here is to predict the next word that is spoken in the meeting. As the input is text, we can apply the transformer with transition matrix model on this dataset where we compute the transition probabilities between parts-of-speech (POS). Lastly, a business use case on a large, proprietary data set is discussed. 

To consider equal time events with $n < N$, we apply different masking strategies for each of the equal time models. In order to apply the LSTM Set model only to time steps with multiple events, (\ref{eq:1}) is implemented in a recurrent layer with a switch function. The cell and hidden states are both initialized as a constant vector and the same attention and LSTM weights are applied at each of the $n$ iterations. For this transformer model, we mask the attention for any missing events because it is possible to have fewer than $N$ co-occurring events at a given time step. When we implement the additional transition matrix, we compute the transition probability between POS on the training set using only the ordered events and apply these probabilities to the rest of the data. 

We compare the classification accuracy across the equal-time models and a baseline model on each dataset. The baseline that we compare against is the averaging model where equal-time events are averaged, such that $\tilde{x_i} = \frac1N\sum_{i\in N} e_i$, and then processed by LSTM. For the transformer equal-time model, we consider both an LSTM model and a transformer model after the equal-time model has been applied to the data. For all other models, we use LSTM for sequence classification. We write each model as U-V where U $\in \{ \text{ds, LSTM, trans, avg}\}$ with ``ds = deep set,'' ``trans = transformer,'' and ``avg = averaging'' represents the underlying equal-time model, and V $\in \{ \text{trans, LSTM}\}$ encodes the model used on $\tilde{x}$. For example, ds-LSTM represents the model with deep set used for equal time and then the resulting ordered sequence is treated by LSTM. 

The transformer equal-time models are compared against the existing models: ds-LSTM, LSTM-LSTM and avg-LSTM. \textcolor{black}{We also consider the trans-trans model as it is an extension of the trans-LSTM model.} In each dataset, we have samples of varying length. Instead of padding the data to a maximum sequence length, we instead group samples by sequence length and train on batches of equal length sequences. On both of the sensor datasets, we merge adjacent sensor readings to create equal-time event sets. As the equal-time events on these datasets are synthetically created, we compare the classification results between the datasets with and without equal-time events. All models are implemented in Keras and trained on a single GPU card. For the business use case and sensor datasets, we report the F1-score as they are classification tasks. For the next word prediction task on the meeting transcript data we report perplexity.

\subsection{Environmental Sensors} 
\begin{figure}[H]
\centering
\begin{subfigure}[t]{0.475\textwidth}
\centering
\includegraphics[width=0.8\textwidth]{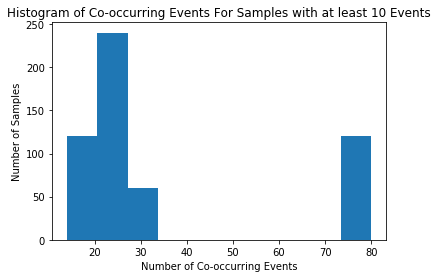}
\end{subfigure}
\begin{subfigure}[t]{0.475\textwidth}
\centering
\includegraphics[width=0.8\textwidth]{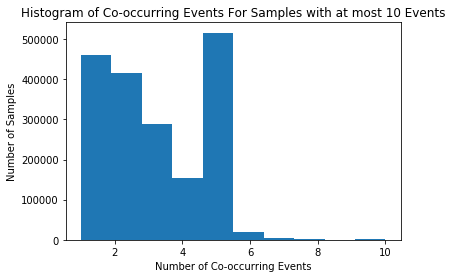}
\end{subfigure}
 \caption{Histogram of Co-occurring Events}
\label{fig:aot-coccur}
\end{figure}

This data set consists of a series of contiguous readings of environmental factors from sensors located around the city of Chicago. The sensor readings include humidity, precipitation, sound in decibels and the amount of light detected. We predict if each of the 52 sensor values will be "large" 12 hours in the future. A sensor value is determined to be large if it is at least one standard deviation over the mean. The data consists of partially ordered sequences with up to 80 co-occurring events and at most 60 time steps in the sequence. The distribution of the number of co-occurring events is shown in Figure \ref{fig:aot-coccur}. As the majority of the time steps have fewer than 10 co-occurring events, the distributions are plotted separately for samples with more than 10 equal time events and samples with fewer than 10 equal time events. The time between sequential readings is computed and if the elapsed time is under a certain threshold, the readings are binned. In this way, multiple sequential readings may be binned together. There are 31 thousand samples in the data set, 70\%  of which comprise the training set. 

We observe in Table \ref{tab:AoT-metrics} that the trans-trans model outperforms all pre-existing, or non-transformer, models. The trans-trans model yields an 11.2\% improvement over the avg-LSTM model and a 10.1\% improvement over the LSTM-LSTM model which is the best existing model for unordered events on this dataset. The improvement over the LSTM-LSTM model is significant with p-value = $1.18\times 10^{-6}$.

\begin{table}[h]
\begin{center}
\caption{Environmental Sensor Performance Metrics (F1 score)}

\resizebox{0.3\textwidth}{!}{\begin{tabular}{l r r}
\toprule
Model & Average & Std Dev \\
\midrule
avg-LSTM & 0.883 & 0.006 \\
ds-LSTM & 0.892 & 0.006 \\
LSTM-LSTM & 0.899 & 0.006 \\
trans-LSTM & 0.908  & 0.004 \\
trans-trans & \textbf{0.983} &  \textbf{0.001} \\
\bottomrule
\end{tabular}}
\label{tab:AoT-metrics}
\end{center}
\end{table}

As this data is ordered and synthetic equal-time events are created by binning, we can train an LSTM model and a transformer model on the ordered data. On this dataset, the LSTM model returns a test F1-score of 0.930 and the transformer model returns a test F1-score of 0.993. The models trained on the ordered data outperform the models trained on the partially ordered data. This is consistent with our expectations as information is lost when equal-time events are created. It is noteworthy that the trans-trans equal-time model (on the partially ordered data) outperforms the LSTM model on the totally ordered data.

\subsection{Household Power} 
\begin{figure}[H]
\centering
\begin{subfigure}[t]{.3\textwidth}
\centering
\caption{Household Power Performance Metrics (F1)}
\resizebox{\textwidth}{!}{\begin{tabular}{l r r}
\toprule
Model & Average & Std Dev \\
\midrule
avg-LSTM & 0.533 & 0.025 \\
ds-LSTM & 0.604 & 0.021 \\
LSTM-LSTM & 0.472 & 0.037 \\
trans-LSTM & 0.591 & \textbf{0.011} \\
trans-trans & \textbf{0.649} & 0.027 \\
\bottomrule
\end{tabular}}

\label{tab:AMP-metrics}
\end{subfigure}
\begin{subfigure}[t]{0.66\textwidth}
 \caption{Histogram of Number of Equal Time Events}
\centering
\includegraphics[width=0.75\textwidth]{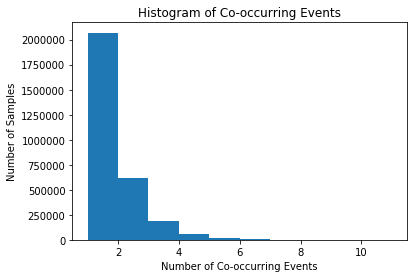}
  \label{fig:amp-eq-time}
\end{subfigure}
\end{figure}

This dataset consists of a sequence of utility readings from a residential household over a two year period. These measurements include water, natural gas and power readings. We use 11 power-related measurements on the household level as features and consider sequences with at most 60 time steps. As with the environmental sensor dataset, readings are binned if the elapsed time between sequential readings is under a set threshold. At a given time step, there are at most 11 co-occurring events as shown in Figure \ref{fig:amp-eq-time}. We use these household power measurements to determine if five appliances are turned on or off (it is a 5-class problem). These appliances are the furnace, heat pump, wall oven, clothes washer and clothes dryer and we threshold the power usage for each appliance in order to label each appliance as on or off. An appliance is classified as on if the power usage is over one standard deviation above the mean. For each sequence, a prediction is made for each appliance and the F1-scores for each class are averaged. There are 20 thousand samples in the data set 70\% of which comprises the training set. 

We observe in Table \ref{tab:AMP-metrics} that the trans-trans model outperforms all pre-existing, or non-transformer, models. The trans-trans model yields a 21.7\% improvement over the avg-LSTM model and a 7.4\% improvement over the ds-LSTM model which is the best pre-existing model for unordered events on this dataset. The improvement over the ds-LSTM model is significant with p-value = 0.02. It is interesting that while the trans-trans model consistently outperforms all pre-existing models on both sensor datasets, the best performing non-transformer model varies between the datasets. This suggests that the non-transformer  equal-time models lack consistency. 

As the utilities readings are ordered and equal-time events are created by binning, we can compare the F1-scores for the equal-time models against the F1-scores for a standard LSTM model and a transformer model on the ordered dataset. On the ordered dataset, the LSTM model returns a test F1-score of 0.653 and the transformer model returns a test F1-score of 0.665. The models trained on the ordered datasets outperform the equal-time models as expected. As we observed on the sensor dataset, the trans-trans model performance is the most similar to the model performance on the ordered data. 

\begin{table}[H]
\parbox{.45\linewidth}{
\centering
\caption{Power Dataset with More Equal-Time Events}
\label{tab:AMP-metrics-more}
\resizebox{0.35\textwidth}{!}{\begin{tabular}{l r r}
\toprule
Model & Average & Std Dev \\
\midrule
avg-LSTM & 0.550 & \textbf{0.017}\\
ds-LSTM & 0.515& 0.034\\
LSTM-LSTM & 0.503 & 0.034 \\
trans-LSTM & 0.522 & \textbf{0.045} \\
trans-trans & \textbf{0.557} & \textbf{0.017}\\
\bottomrule
\end{tabular}}}
\hfill
\parbox{.45\linewidth}{
\centering

\caption{Power Dataset with Fewer Equal Time Events}
\label{tab:AMP-metrics-less}
\resizebox{0.35\textwidth}{!}{\begin{tabular}{l r r}
\toprule
Model & Average & Std Dev \\
\midrule
avg-LSTM & 0.573 & 0.026\\
ds-LSTM & 0.591 & 0.025\\
LSTM-LSTM & 0.534 & \textbf{0.013} \\
trans-LSTM & 0.554 & 0.026 \\
trans-trans & \textbf{0.656} & 0.019\\
\bottomrule
\end{tabular}}}
\end{table}

To examine the effect of the number of equal-time events on model performance, we consider a dataset with more equal-time events and a dataset with fewer equal-time events. These datasets have at most 24 co-occurring events and 5 co-occurring events, respectively. We observe in Table ~\ref{tab:AMP-metrics-more} that when there are more equal-time events, the scores for equal-time models are on the whole lower which is expected. The trans-trans model returns the highest average F1-score in all datasets. In addition, the differences in F1-score for the power dataset with more events between the best performing existing model, avg-LSTM, and the trans-trans model are not statistically significant. The trans-trans model outperforms the best performing existing model, ds-LSTM, and the difference is significant with p-value=0.002 in Table ~\ref{tab:AMP-metrics-less}. Therefore, we observe that the trans-trans model outperforms the best pre-existing model across different levels of equal-time events.

\subsection{Meeting Transcripts} 
\begin{figure}[h]
\begin{subfigure}[t]{0.4\textwidth}
\begin{center}
\caption{Meeting Transcript Performance Metrics}
\resizebox{\textwidth}{!}{\begin{tabular}{c r r }
\toprule
Model & Average & Std. Dev \\
\midrule
trans-trans & 1036.40 & 114.93 \\
trans-trans with transition matrix & \textbf{951.47} & \textbf{40.12} \\
\bottomrule
\end{tabular}}

\label{tab:AMI-metrics}
\end{center}
\end{subfigure}
\begin{subfigure}[t]{0.6\textwidth}
 \caption{Histogram of Number of Equal Time Events}
\centering
\includegraphics[width=0.75\textwidth]{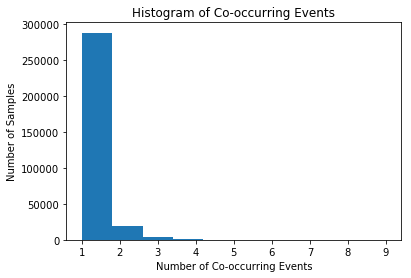}
  \label{fig:ami-eq-time}
\end{subfigure}
\end{figure}

This dataset consists of transcriptions of audio meeting recordings with four speakers in the room from the Augmented Multi-party Interaction (AMI) dataset. These recordings contain both scripted and unscripted meetings. Of the words that are uttered in the recording, 20.2\% of words occur on co-occurring events, which make up 7.4\% of all timestamps. At a given timestamp, we have at most 9 co-occurring words as shown in Figure \ref{fig:ami-eq-time}.  As this is a language dataset, we consider the next word prediction task. There are 3,586 words in the vocabulary, sequences are on average 35 time steps long and there are 17 thousand samples with 70\% in the training set. Words are embedded using a pretrained BERT model \cite{bert}. For the transformer model with transition matrix, we use the default NLTK part-of-speech tagger to compute the part-of-speech for each word in the transcript \cite{nltk}. We report test perplexity for each of the models.

As a baseline comparison, we compare the trans-trans with transition matrix model against both the trans-trans model and a pre-trained GPT-2 model. In order to apply the GPT-2 model, we randomly order equal-time events and then use the XL pre-trained GPT-2 weights to infer the next word  \cite{gpt-2}. This model returns a test perplexity of 2,226.63. We observe in Table \ref{tab:AMI-metrics} that both transformer models significantly outperform the GPT-2 model. We only benchmark the transition matrix model against the trans-trans model as we want to determine if the transition matrix input improves on the transformer equal-time representation. Furthermore, when we consider the average test perplexity across five runs with different initial seeds, we observe that the transformer model with transition matrix input results in an 8.21\% decrease in test perplexity over the trans-trans model. 

Therefore, we observe that for the dataset which is inherently unordered, when a random ordering is chosen for equal-time events, it results in a higher perplexity than when an equal-time model is applied. This is the opposite of what is observed in the two sensor datasets, which makes sense as those datasets are inherently ordered and synthetic equal-time events were generated through binning. We also show that the transition matrix input to the self-attention calculation improves model performance. However, the transition matrix input can be considered only for datasets such as this one where the ordering of different event types is known and is meaningful. 

\subsection{Business Use Case} 

This dataset consists of a prediction task on a partially ordered sequence dataset. Sequences are at most 100 time steps long and at a given time step we have at most 20 co-occurring events. There are on the order of 1 million samples, with about 70\% in the training set. Comparing the results of each of the proposed models against the baseline averaging model in Table \ref{tab:perform-metrics}, we observe that the avg-LSTM model outperforms the proposed equal-time models. When we ensemble across all of the proposed models, we report a test F1-score of 0.22, which outperforms the avg-LSTM model. We do not observe the same improvement in behavior when we ensemble across multiple runs of the averaging models. On this dataset, a non-deep learning model with hand selected features returns an F1-score of 0.22. While these deep learning equal-time models do not outperform a non-deep learning model, we achieve the same F1-score without feature selection. 
The co-occurrence of events in a superficial sense is arising from the coarse granularity of the time stamps. The transformer approaches not being able to realize gains similar to the ones observed in the other use cases triggered a closer examination of the data. The conclusion was that time stamps for a significant subset of the events represent merely start or end points of longer time periods during which the event in question actually occurred. This is a major difference to the other three use cases. Treating these time stamp as actual event occurrence introduces misleading information. We suspect this to contribute to the performance gap seen in the tranformer models.
Consequently, in future work, we will need to consider co-occurence of such events across multiple time steps in a probabilistic sense. 

\begin{table}[h!]
\begin{center}
\caption{Performance Metrics}
\resizebox{0.45\textwidth}{!}{\begin{tabular}{l r r}
\toprule
 Model & Validation F1 & Test F1\\
\midrule
avg-LSTM & 0.245 & 0.217 \\
ds-LSTM & 0.227 & 0.207 \\
LSTM-LSTM & 0.224 & 0.207 \\
trans-LSTM & 0.229 & 0.212 \\
trans-trans & 0.219 & 0.204 \\
\midrule
ensemble & 0.237 & 0.220 \\
\bottomrule
\end{tabular}}

\label{tab:perform-metrics}
\end{center}
\end{table}

\section{Conclusion}
We have presented several techniques for classification of partially ordered sequence data. Models were evaluated on four data sets, where it was observed that the transformer model for equal-time events outperforms models that incorporate existing order-invariant techniques. On data sets where the data is inherently ordered and synthetic events are generated by binning, models trained on the ordered sequences outperform the equal-time models. On a language dataset with unordered data, the equal-time models outperform a language model with randomly ordered equal-time events. Finally, we provide evidence on the meeting transcript data that transition probabilities between events can be used to further improve model performance.

\begin{footnotesize}
\centering\bibliography{equal-time}
\end{footnotesize}
\end{document}